\documentclass{article}
\usepackage{corl_2017}

\usepackage{hyperref}  

\usepackage{amsmath}
\usepackage{amsfonts}
\usepackage{mathtools}

\usepackage{booktabs}

\usepackage{algorithm}
\usepackage{algorithmic}

\usepackage{color}

\usepackage{graphicx}
\usepackage{caption}
\usepackage{subcaption}
\usepackage{wrapfig}

\usepackage{siunitx}

\long\def\ignorethis#1{}

\def\equationautorefname~#1\null{Equation~(#1)\null}

\renewcommand{\eqref}[1]{\autoref{#1}}

\newcommand{\action}{\mathbf{a}}
\newcommand{\state}{\mathbf{x}}

\newcommand{\pixel}{d} 
\newcommand{\goal}{g}

\usepackage{makecell}

\usepackage{authblk}

\begin{document}

\newcommand{\specificthanks}[1]{{#1}}

\title{Self-Supervised Visual Planning with Temporal Skip Connections}

\author[1, 2]{Frederik Ebert}
\author[1]{Chelsea Finn}
\author[1]{Alex X. Lee}
\author[1]{Sergey Levine}

\affil[1]{\footnotesize Department of Electrical Engineering and Computer Sciences, UC Berkeley, United States}
\affil[2]{\footnotesize Department of Computer Science, Technical University of Munich, Germany}

\affil[ ]{\texttt{\{febert,cbfinn,alexlee\_gk,svlevine\}@berkeley.edu}}

\maketitle

\begin{abstract}
In order to autonomously learn wide repertoires of complex skills, robots must be able to learn from their own autonomously collected data, without human supervision. One learning signal that is always available for autonomously collected data is prediction. If a robot can learn to predict the future, it can use this predictive model to take actions to produce desired outcomes, such as moving an object to a particular location. However, in complex open-world scenarios, designing a representation for prediction is difficult. In this work, we instead aim to enable self-supervised robot learning through direct video prediction: instead of attempting to design a good representation, we directly predict what the robot will see next, and then use this model to achieve desired goals. A key challenge in video prediction for robotic manipulation is handling complex spatial arrangements such as occlusions. To that end, we introduce a video prediction model that can keep track of objects through occlusion by incorporating temporal skip-connections. Together with a novel planning criterion and action space formulation, we demonstrate that this model substantially outperforms prior work on video prediction-based control. Our results show manipulation of objects not seen during training, handling multiple objects, and pushing objects around obstructions. These results represent a significant advance in the range and complexity of skills that can be performed entirely with self-supervised robot learning. 
\end{abstract}


\section{Introduction}


\begin{wrapfigure}{r}{0.3\textwidth}
\vspace{-0.2in}
    \begin{center}
        \includegraphics[width=0.3\textwidth]{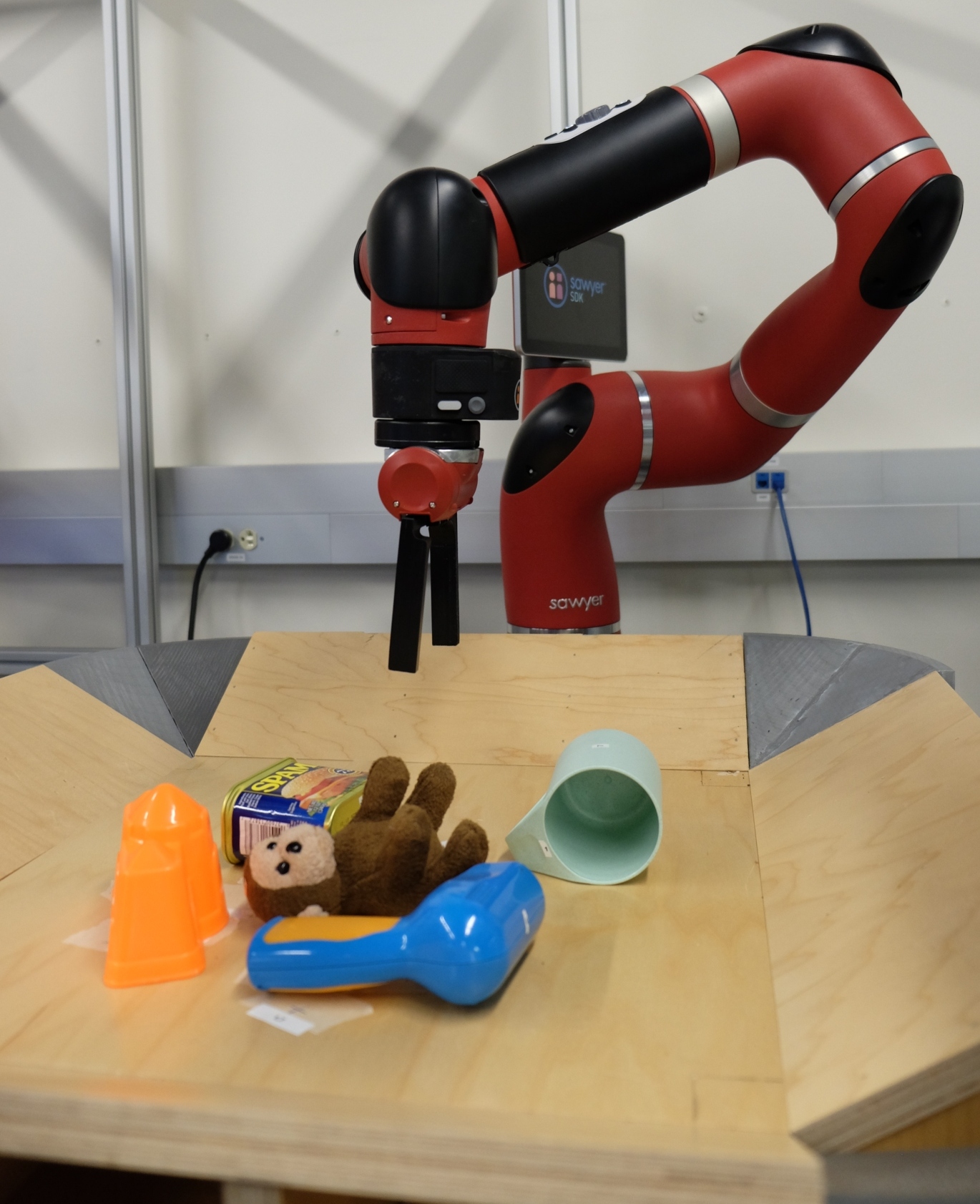}
    \end{center}
    \vspace{-0.1in}
    \caption{The robot learns to move new objects from self-supervised experience.} 
    \label{fig:teaser}
    \vspace{-0.4cm}
\end{wrapfigure}

A key bottleneck in enabling robots to autonomously learn a wide range of skills is the need for human involvement, for example, through hand-specifying reward functions, providing demonstrations, or resetting and arranging the environment for episodic learning.
Learning action-conditioned predictive models of the environment is a promising approach for self-supervised robot learning as it allows the robot to learn entirely on its own from autonomously collected data. In order for a robot to be able to predict what will happen in response to its actions, it needs a representation of the environment that is suitable for prediction. In complex, open-world environments, it is difficult to construct a concise and sufficient representation for prediction. At a high level, the number and types of objects in the scene might change substantially trial to trial, making it impossible to provide a single fixed-size representation. At a low level, the type of information that is important about each object might change from object to object: the motion of a rigid box might depend only on its shape and friction coefficient, while the motion of a deformable stuffed animal might depend on a variety of other factors which are hard to determine by hand. Instead of engineering representations for different object types and object scenes, we can instead directly predict the robot's sensory observations. The rationale behind this is that, if the robot is able to predict future observations, it has acquired a sufficient understanding of the environment that it can leverage for planning actions.

A major challenge in directly predicting visual observations for robotic control lies in deducing the spatial arrangement of objects from ambiguous observations. For example, when one object passes in front of another, the robot must remember that the occluded object still persists. In human visual perception, this is referred to as object permanence, and is known to take several months to emerge in infants~\cite{permanence}. When the robot is commanded to manipulate an object which can become occluded during a manipulation, it must be able to accurately predict how that object will respond to an occlusion.
In this work, we propose a visual predictive model for robotic control that can reason about spatial arrangements of objects in 3D, while using only monocular images and without providing any form of camera calibration. Previous learning methods have proposed to explicitly model object motion in image-space~\cite{foresight,dynamic_filter_networks} or to explicitly model 3D motion using point cloud measurements from depth cameras~\cite{se3}, and lacked the capability to maintain information about objects which are occluded during the predicted motion. We propose a simple model that does not require explicitly deducing full 3D structure in the scene, but does provide effective handling of occlusions by storing the appearance of occluded objects in memory. Furthermore, unlike prior prediction methods~\cite{se3,optical_flow_pred} our model does not require any additional, external systems for providing point-to-point correspondences between frames of different time-steps.

The technical contribution of this work is three-fold. First, we present a video prediction method that can more accurately maintain object permanence through occlusions, by incorporating temporal skip connections. Second, we propose a planning objective for control through video prediction that leads to significantly improved long-term planning performance, including planning through occlusions, when compared to prior work~\cite{foresight}. Finally, we propose a mechanism for planning with both discrete and continuous actions with video prediction models. Our evaluation demonstrations that these components can be combined to enable a learned video prediction model to perform a range of real-world pushing tasks. Our experiments include manipulation of previously unseen objects, handling multiple objects, pushing objects around obstructions, and moving the arm around and over other obstacle-objects, representing a significant advance in the range and complexity of skills that can be acquired through entirely self-supervised learning.

\section{Related Work}

\paragraph{Large-Scale, Self-Supervised Robotic Learning.}

Large-scale robotic data collection has been explored in a number of recent works. \citet{ot-aaib-15} proposed to collect object scans using multiple robots.
Several prior works have focused on autonomous data collection for individual skills, such as grasping~\cite{lerrel,google_handeye} or obstacle avoidance~\cite{greg_kahn_uncertainty,crashing}. In contrast to these methods, our approach learns predictive models that can be used to perform a variety of manipulations, and does not require a success measure or reward function during data collection. Several prior approaches have also sought to learn inverse or forward models from raw sensory data without any supervision~\cite{pulkit,foresight}.
While these methods demonstrated effective generalization to new objects, they were limited in the complexity of tasks and time-scale at which these tasks could be performed. The method proposed by~\citet{pulkit} was able to plan single pokes, and then greedily execute multiple pokes in sequence. The method of~\citet{foresight} performed long-horizon planning, but was only effective for short motions. In our comparisons to this method, we demonstrate a substantial improvement in the length and complexity of manipulations that can be performed with our models.

\paragraph{Sensory Prediction Models.}

Action-conditioned video prediction has been explored in the context of synthetic video game images~\cite{atarioh,recurrentsimulators} and robotic manipulation~\cite{bootsetal,finn_nips,video_pixel_networks}, and video prediction without actions has been studied for unstructured videos~\cite{beyond_mse,convlstm,vondrick} and driving~\cite{prednet,dynamic_filter_networks}. 

Several works have sought to use more complex distributions for future images, for example by using autoregressive models~\cite{video_pixel_networks,scott_reed}. While this often produces sharp predictions, the resulting models are extremely demanding computationally, and have not been applied to real-world robotic control. In this work, we extend video prediction methods that are based on predicting a transformation from the previous image~\cite{finn_nips,dynamic_filter_networks}. Prior work has also sought to predict motion directly in 3D, using 3D point clouds obtained from a depth camera~\cite{se3}, requiring point-to-point correspondences over time, which makes it hard to apply to previously unseen objects. Our predictive model is effective for a wide range of real-world object manipulations and does not require 3D depth sensing or point-to-point correspondences between frames.
Prior work has also proposed to plan through learned models via differentiation, though not with visual inputs~\cite{deep_mpc}. We instead use a stochastic, sampling-based planning method~\cite{cem-rk-13,foresight}, which we extend to handle a mixture of continuous and discrete actions.

\section{Preliminaries}

In this section, we define our image-based robotic control problem, present a formulation of visual model predictive control (visual MPC) over pixel motion, and summarize prior video prediction models based on image transformation. 

\subsection{Visual Model Predictive Control}
\label{sec:vmpc}

Our visual MPC problem formulation follows the problem statement outlined in prior work~\cite{foresight}. We assume that the user defines a goal for the robot in terms of pixel motion: given an image from the robot's camera, the user can choose one or more pixels in the image, and choose a destination where each pixel should be moved. For example, the user might select a pixel on an object and ask the robot to move it 10 cm to the left. Formally, the user specifies $P$ source pixel locations $\pixel_0^{(1)}, \dots, \pixel_0^{(P)}$ in the initial image $I_0$, and $P$ goal locations $\goal^{(1)}, \dots, \goal^{(P)}$. The source and goal pixel locations are denoted by the coordinates $(x_d^{(i)}, y_d^{(i)})$ and $(x_g^{(i)}, y_g^{(i)})$. Given a goal, the robot plans for a sequence of actions $\action_{1:T}$ over $T$ time steps, where $T$ is the planning horizon. The problem is formulated as the minimization of a cost function $c$ which depends on the predicted pixel positions $d_t^{(j)}$. The planner makes use of a learned model that predicts pixel motion. Given a distribution over pixel positions $P_{t_0, d^{(i)}}\in\mathbb{R}^{H\times W}, \sum_{H,W} P_{t_0, d^{(i)}} = 1$ at time $t = 0$, the model predicts distributions over its positions $P_{t, d^{(i)}}$ at time $t \in \{ 1, \dots, T \}$.
To achieve the best results with imperfect models, the actions can be iteratively replanned at each real-world time step $\tau \in \{0,...,\tau_{max}\}$ following the framework of model-predictive control (MPC): at each real-world step $\tau$, the model is used to plan $T$ steps into the future, and the first action of the plan is executed.
At the first real-world time step $\tau=0$, the distribution $P_{t=0, d^{(i)}} $ is initialized as 1 at the location of the designated pixel and zero elsewhere. In subsequent steps ($\tau > 0$), the 1-step ahead prediction of the previous step is used to initialize $P_{t=0,d^{(i)}}$. The cost $c$ is a function of $P_{t,d^{(i)}}$ at all predicted time steps $0...T$. Prior work proposed to use the negative log-probability of placing each pixel at its goal location as the cost~\cite{foresight}. We will discuss in Section~\ref{sec:mpc_cost} how performance can be improved substantially by instead using expected distances. Planning is performed using the cross-entropy method (CEM), a gradient-free optimization procedure that consists of iteratively resampling action sequences and refitting Gaussian distributions to the actions with the best predicted cost. Further details can be found in prior work~\cite{foresight}.

\subsection{Video Prediction via Pixel Transformations}
\label{sec:model}
Visual MPC requires a model that can effectively predict the motion of the selected pixels $\pixel_0^{(1)}, \dots, \pixel_0^{(P)}$ up to $T$ steps into the future.
In this work, we extend the model proposed in \cite{finn_nips}, where this flow prediction capability emerges implicitly, and therefore no external pixel motion supervision is required. Future images are generated by transforming past observations. The model uses stacked convolutional LSTMs that predict a collection of pixel transformations at each time step, with corresponding composition masks. In this model, the previous image $I_t$ is transformed by $N$ separate transformations, and all of the transformed images $\tilde{I}_t^{(i)}$ are composited together according to weights obtained from the predicted masks. Intuitively, each transformation corresponds to the motion of a different object in the scene, and each mask corresponds to the spatial extents of that object. Let us denote the $N$ transformed images as $\tilde{I}_t^{(1)}, ..., \tilde{I}_t^{(N)}$, and the predicted masks as $\mathbf{M}_1, ...\mathbf{M}_N$, where each 1-channel mask is the same size as the image. The next image prediction is then computed by compositing the images together using the masks: $\hat{I}_{t+1} = \sum_{i=1}^N \tilde{I}_t^{(i)} \mathbf{M}_i$. To predict multiple time steps into the future, the model is applied recursively. These transformations can be represented as convolutions, where each pixel in the transformed image is formed by applying a convolution kernel to the previous one. This method can represent a wide range of local transformations. When these convolution kernels are normalized, they can be interpreted as transition probabilities, allowing us to make probabilistic predictions about future locations of individual pixels. To predict the future positions of the designated pixels $d^{(i)}$, the same transformations which are used for the images are applied to $P_{t,d^{(i)}}$ such that $P_{t+1,d^{(i)}} = \frac{1}{P_s}\sum_{i=1}^N \tilde P^{(i)}_{t,d^{(i)}} \mathbf{M}_i $, where $\frac{1}{P_s}$ is a normalizing constant to ensure that $P_{t+1,d^{(i)}}$ adds up to 1 over the spatial dimension of the image. Since this prior predictive model outputs a single image at each time step, it is unable to track pixel positions through occlusions. Therefore, this model is only suitable for planning motions where the user-selected pixels are not occluded during the manipulation, which restricts its use in cluttered environments or with multiple selected pixels. In the next section, we discuss our proposed occlusion-aware model, which lifts this limitation by employing temporal skip connections.

\section{Skip Connection Neural Advection Model}
\label{sec:occlusion_model}
\begin{figure}
    \centering
    \begin{minipage}{.3\textwidth}
        \centering
        \includegraphics[width=0.9\textwidth]{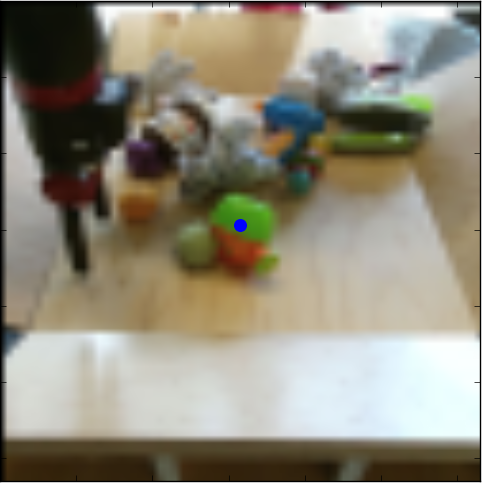}
        \caption{The blue dot indicates the designated pixel}
        \label{fig:desig_pix_bluedot}
    \end{minipage}
    \hfill
    \begin{minipage}{.65\textwidth}
        \centering
        \includegraphics[width=\textwidth]{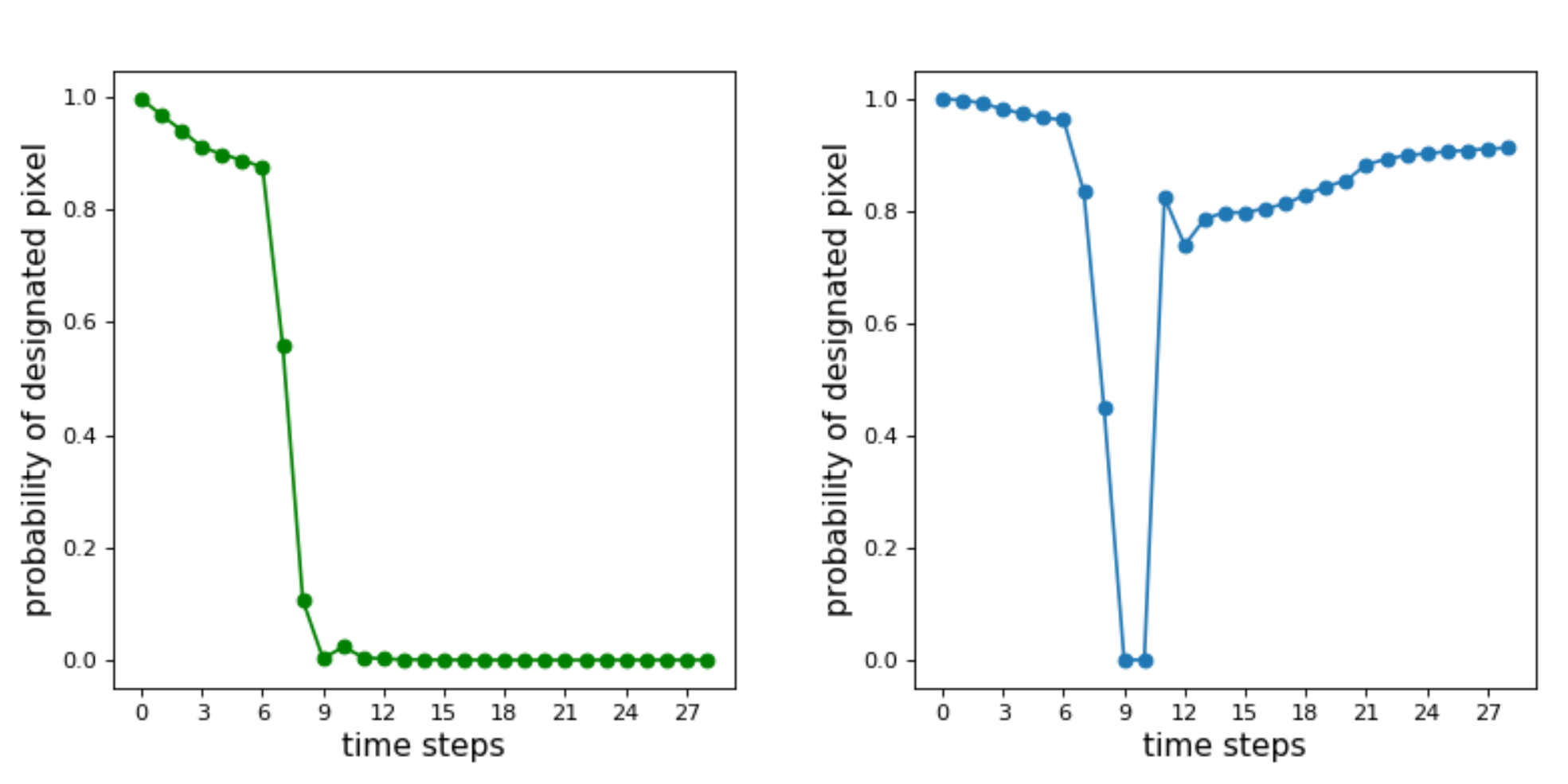}
        \caption{Predicted probability $P_{d^{(0)}}(t)$ of the designated pixel being at the location of the blue dot indicated in \autoref{fig:desig_pix_bluedot} for the DNA model (left) and the SNA model (right).}        \label{fig:pix_reqppear_graph}
    \end{minipage}
\end{figure}

To enable effective tracking of objects through occlusions, we propose an extension to the dynamic neural advection (DNA) model~\cite{finn_nips} that incorporates temporal skip connections. This model uses a similar multilayer convolutional LSTM structure; however, the transformations are now conditioned on a context of previous images. In the most generic version, this involves conditioning the transformation at time $t$ on all of the previous images $I_0,...I_{t}$, though in practice we found that a greatly simplified version of this model performed just as well in practice. We will therefore first present the generic model, and then describe the practical simplifications. We refer to this model as the skip connection neural advection model (SNA), since it handles occlusions by copying pixels from prior images in the history such that when a pixel is occluded (e.g., by the robot arm or by another object) it can still reappear later in the sequence. When predicting the next image $\hat{I}_{t+1}$, the generic SNA model transforms each image in the history according to a different transformation and with different masks to produce $\hat{I}_{t+1}$ (see \autoref{fig:general_model} in the appendix):
\begin{equation}
\hat{I}_{t+1} =  \sum_{j=t-T}^{t} \sum_{i=1}^{N} \mathbf{M}_{i,j} \tilde{I}_{j}^{(i)}
\label{eqn:general_model}
\end{equation}
In the case where $t < T$, negative values of $j$ simply reuse the first image in the sequence. This generic formulation can be computationally expensive, since the number of masks and transformations scales with $T \times N$. A more tractable model, which we found works comparably well in practice in our robotic manipulation setting, assumes that occluded objects are typically static throughout the prediction horizon. This assumption allows us to dispense the intermediate transformations and only provide a skip connection from the very first image in the sequence, which is also the only real image, since all of the subsequent images are predicted by the model:
\begin{equation}
\hat{I}_{t+1} =  I_0 \mathbf{M}_{N+1} +  \sum_{i=1}^{N} \tilde{I}_t^{(i)} \mathbf{M}_i
\label{eqn:simplemodel}
\end{equation}
This model only needs to output $N+1$ masks. We observed similar prediction performance when using a transformed initial image $\tilde{I}_0$ in place of $I_0$, and therefore used the simplified model in \autoref{eqn:simplemodel} in all of our experiments. We provide an example of the model recovering from occlusion in \autoref{fig:pix_reappear}. In the figure, the arm moves in front of the designated pixel, marked in blue in \autoref{fig:desig_pix_bluedot}. The graphs in \autoref{fig:pix_reqppear_graph} show the predicted probability of the designated pixel, which is stationary during the entire motion, being at its original position at each step. Precisely when the arm occludes the designated pixel, the pixel's probability of being at this point decreases. This indicates that the model is `unsure' where this pixel is. When the arm unoccludes the designated pixel, it should become visible again, and the probability of the designated pixel being at its original position should go up. In the case of the DNA model and its variants~\cite{finn_nips}, the probability mass does not increase after the object reappears. This is because the DNA model cannot recover information about object that it has `overwritten' during its predictions.
Consequently, for these models, the probability stays low and \emph{moves with the arm}, which causes the model to believe that the occluded pixel also moves with the arm. We identified this as one of the major causes of planning failure when using the prior models. By contrast, in the case of our SNA model, the probability of the correct pixel position increases again rapidly right after the arm unoccludes the object. Furthermore, the probability of the unoccluded object's position becomes increasingly sharp at its original position as the arm moves further away.

\begin{figure}
    \centering
    \includegraphics[width=\textwidth]{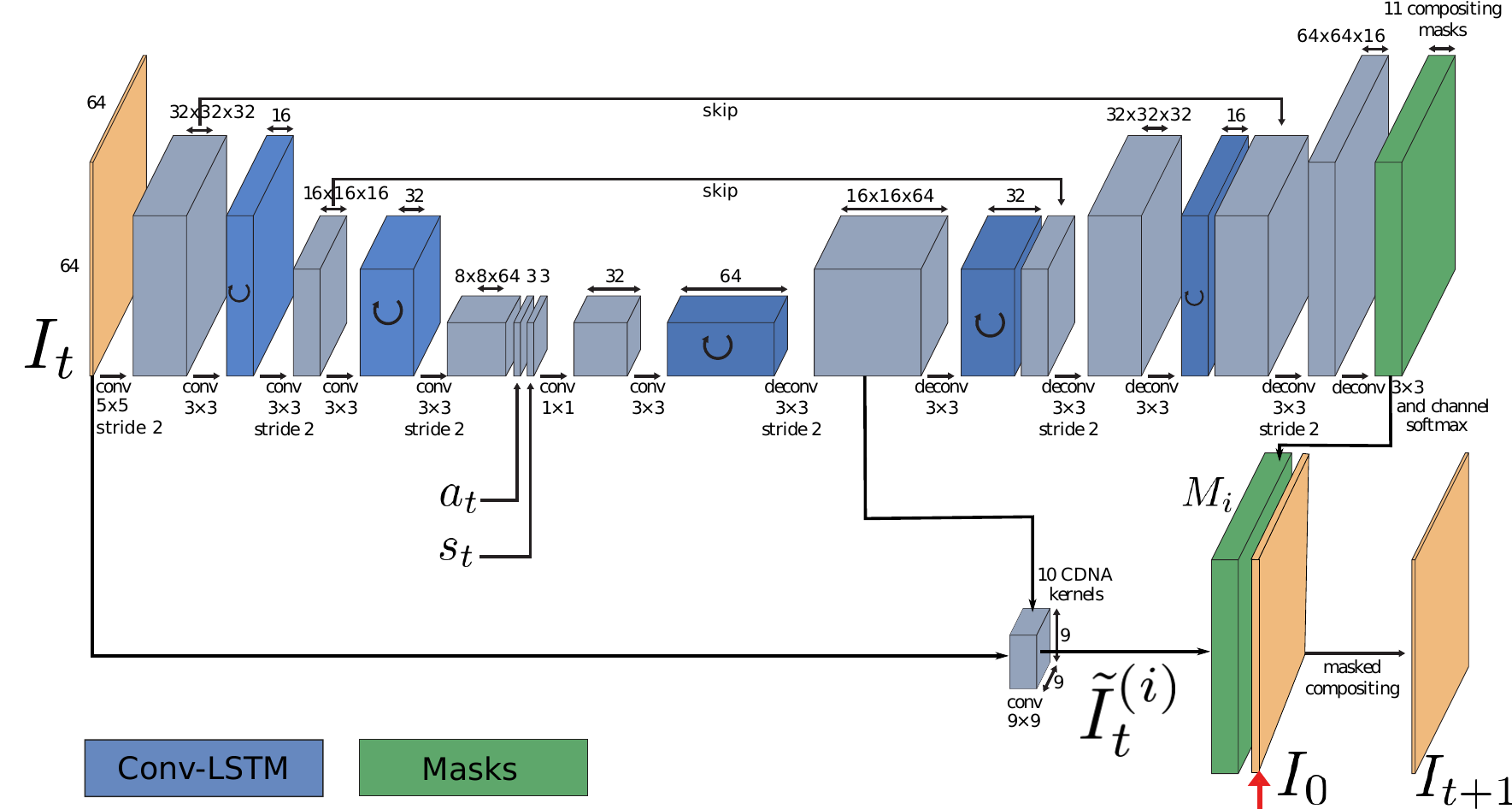}
    \caption{Simplified SNA model based on \autoref{eqn:simplemodel}. The red arrow indicates where the image from the first time step $I_0$ is concatenated with the transformed images $\tilde{I}^{(i)}_t$ multiplying each channel with a separate mask to produce the predicted frame for step $t+1$.}      \label{fig:occlusion_model}
\end{figure}

\begin{figure}
    \centering
    \begin{subfigure}{0.9\textwidth}
    \centering
        \includegraphics[width=0.8\linewidth]{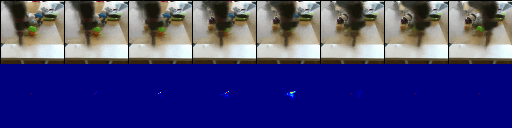}
        \caption{Skip connection neural advection (SNA) does not erase or move objects in the background}
        \label{fig:Ng1}
    \end{subfigure}
    \begin{subfigure}{0.9\textwidth}
    \centering
        \includegraphics[width=0.8\linewidth]{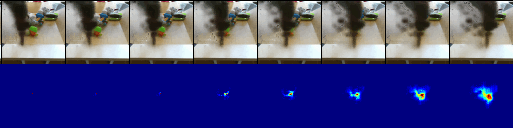}
        \caption{Standard DNA \cite{foresight} exhibits undesirable movement of the distribution $P_{d^{(0)}}(t)$ and erases the background}
        \label{fig:pix_reappear}
    \end{subfigure}
    \caption{
    Top rows: Predicted images of arm moving \textit{in front of} green object with designated pixel (as indicated in \autoref{fig:desig_pix_bluedot}). 
    Bottom rows: Predicted probability distributions $P_{d^{(0)}}(t)$ of designated pixel obtained by repeatedly applying transformations.}
    \label{fig:pix_reappear}
\end{figure}

\section{Visual MPC with Pixel Distance Costs}
\label{sec:mpc_cost}

The choice of objective function for visual MPC has a large impact on the performance of the method. Intuitively, as the model's predictions get more uncertain further into the future, the planner relies more and more on the cost function to provide a reasonable estimate of the overall distance to the goal. Prior work used the probability of the chosen pixel(s) reaching their goal position(s) after $T$ time steps, as discussed in Section~\ref{sec:vmpc}. When the trajectories needed to reach the goal are long, the objective
 provides relatively little information about the progress towards the goal, since most of the predicted probabilities will have a value close to zero at the goal-pixel. In this work, we propose a cost function that still makes use of the uncertainty estimates about the pixel position, but provides a substantially smoother planning objective, resulting in improved performance for more complex, longer-horizon tasks. A straightforward choice of smooth cost function in deterministic settings is the Euclidean distance between the current and desired location of the desired pixel(s), given by $\| \pixel_{t^\prime} - \goal \|_2$. Since our video prediction model produces a distribution over the pixel location at each time step, we can use the expected value of this distance as a cost function, summed over the entire horizon $T$:
\begin{align}
    c_{t+1:t+T}(\goal) &= c(I_{t-1:t}, \state_{t-1:t}, \action_{t:t+T}, \pixel_t, \goal) \nonumber\\
    &= \sum_{t^\prime = t+1, \dots, t+T} \mathbb{E}_{\pixel_{t^\prime} \sim P_{t^\prime,d^{(i)}}} \left[\|\pixel_{t^\prime} - \goal\|_2\right]  
      \label{eq:cost}
  \end{align}
\noindent where the expectation can be computed by summing over all of the positions in each predicted image, and the cost corresponds to a (element-wise) Hadamard product of the pixel location probability map and the distances between each pixel position and the goal.
This cost function encourages the movement of the designated objects in the right direction for each step of the execution, regardless of whether the $\goal$ position can be reached within $T$ time steps. For multi-objective tasks with multiple designated pixels $d^{(i)}$ the costs are summed to together weighting them equally. Although the use of well-shaped cost functions in MPC has been explored extensively in prior work~\cite{Tassa}, the combination of cost shaping and visual MPC has not been studied extensively.

\section{Sampling-Based MPC with Continuous and Discrete Actions}
\label{sec:discrete}

\begin{figure}
    
    \centering
    \vspace{-0.1in}
    \includegraphics[width=1\textwidth]{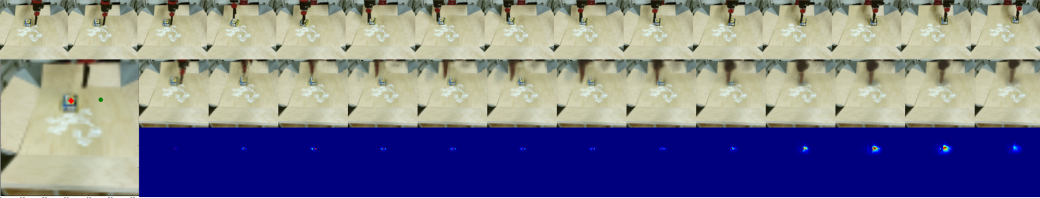}
    \caption{Integrating actions in the vertical direction enables moving over obstacles. This results in more natural and shorter paths. Top row: executed trajectory, left: red diamond indicates designated pixel, green dot indicates goal, 2nd and 3rd row: predictions and probability of designated pixel at time step 3.}
    \label{fig:discrete}
    \vspace{-0.2in}
    
\end{figure}

The choice of action representation for visual MPC has a significant effect on the performance of the resulting controller. The action representation must allow the planner sufficient freedom to maneuver the arm to perform a wide variety of tasks, while constraining sufficiently so as to create a tractable search space. We found that a particularly well-suited action representation for tabletop manipulation can be constructed by combining continuous and discrete actions, in contrast to prior work that used only continuous end-effector motion vectors as actions~\cite{foresight}. The actions consist of the horizontal motion of the end-effector, as well as a discrete ``lift'' action that allows the robot to command the end-effector to lift off the table vertically. The discrete lifting action can take on $N$ values (4 in our implementation) that specify for how many time steps the robot should lift the end-effector off the table. Unlike with continuous vertical motion commands, these discrete commands result in the end-effector staying off the table for multiple time steps even during random data collection. In order to incorporate this hybrid action space into the stochastic CEM-based optimization in visual MPC, we sample real-valued parameters for the discrete action, and then round them to the nearest valid integer to obtain discrete actions. As usual with CEM, we iteratively refit a multivariate Gaussian distribution to the best performing samples (which in our case are chosen to be in the 90$^\text{th}$ percentile of samples), treating the entire action vector as if it was continuous. \autoref{fig:discrete} shows an example of a situation where the discrete vertical motion component of the action is used by our model to lift the gripper over the object in order to push it from the opposite side.

\section{Experiments}

Our experimental evaluation compares the proposed occlusion-aware SNA video prediction model, as well as the improved cost function for planning, with a previously proposed model based on dynamic neural advection (DNA)~\cite{foresight}. We use a Sawyer robot, shown in \autoref{fig:teaser}, to push a variety of objects in a tabletop setting. In the appendix, we include a details on hyperparameters, analysis of sample complexity, and a discussion of robustness and limitations. We evaluate long pushes and multi-objective tasks where one object must be pushed without disturbing another. The supplementary video and links to the code and data are available at \url{https://sites.google.com/view/sna-visual-mpc}


\begin{wrapfigure}{r}{.37\columnwidth}
\vspace{-0.25in}
\centering
        \includegraphics[width=0.30\columnwidth]{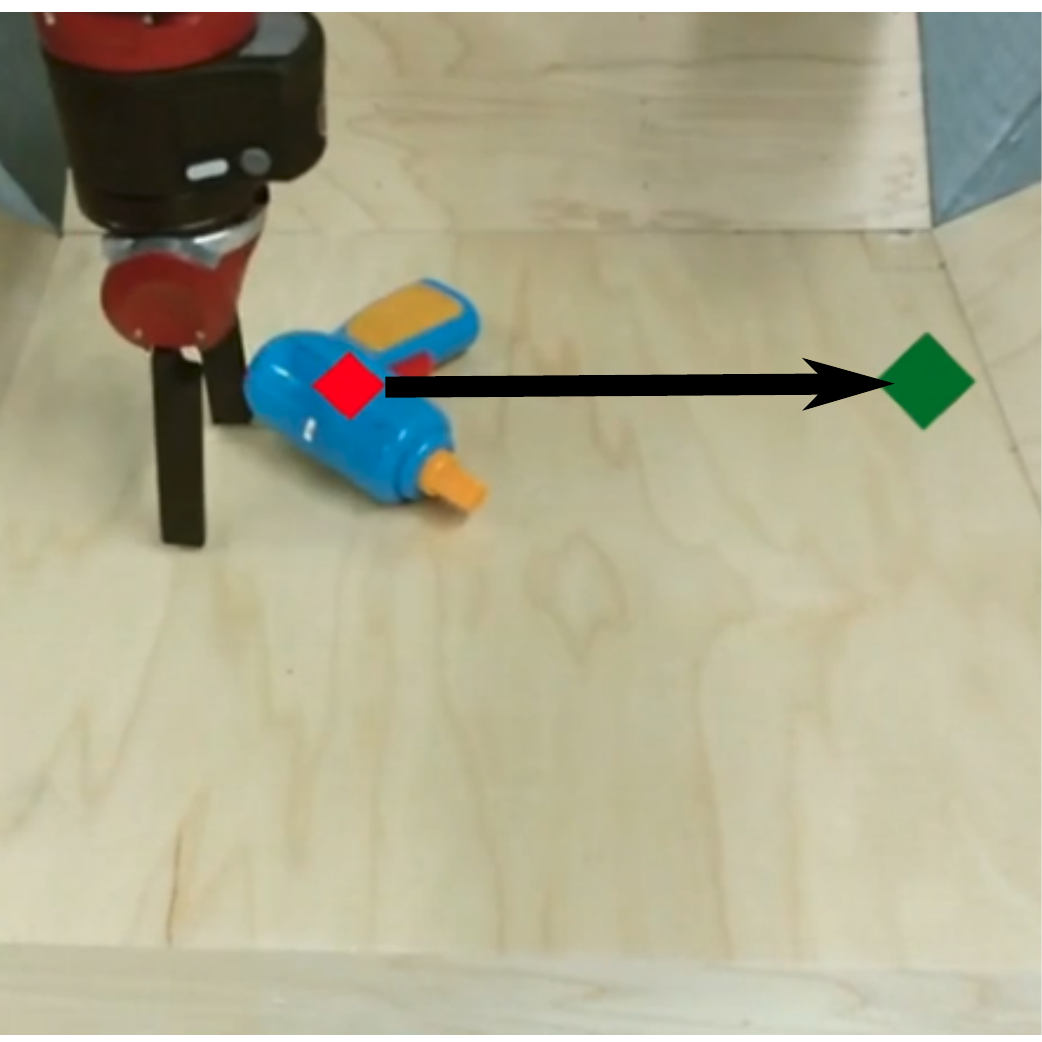}
\caption{
Pushing task. The designated pixel (red diamond) needs to be pushed to the green circle.
\label{fig:long_distance_task}
\vspace{-0.2in}
}
\end{wrapfigure}
\autoref{fig:long_distance_task} shows an example task for the pushing benchmark. We collected 20 trajectories with 3 novel objects and 1 training object. \autoref{table:res_longd} shows the results for the pushing benchmark. The column \textit{distance} refers to the mean distance between the goal pixel and the designated pixel at the final time-step. The column \textit{improvement} indicates how much the designated pixel of the objects could be moved closer to their goal (or further away for negative values) compared to the starting location. The true locations of the designated pixels after pushing were annotated by a human. 

The results in \autoref{table:res_longd} show that our proposed planning cost in \autoref{eq:cost} substantially outperforms the planning cost used in prior work~\cite{foresight}. The performance of the SNA model in these experiments is comparable to the prior DNA model~\cite{foresight} when both use the new planning cost, since this task does not involve any occlusions. Although the DNA model has a slightly better mean distance compared to SNA, it is well within the standard deviation, suggesting that the difference is not significant.

\begin{figure}
\centering
   \includegraphics[width=1\linewidth]{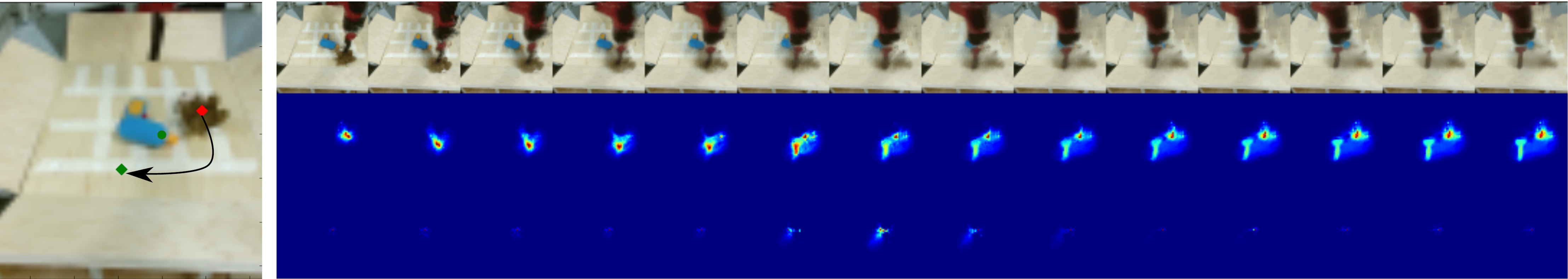}
\caption{Left: Task setup with green dot marking the obstacle. Right, first row: the predicted frames generated by SNA. Second row: the probability distribution of the designated pixel on the \textit{moving} object (brown stuffed animal). Note that part of the distribution shifts down and left, which is the indicated goal. Third row: the probability distribution of the designated pixel on the obstacle-object (blue power drill). Although the distribution increases in entropy during the occlusion (in the middle), it then recovers and remains on its original position.
\label{fig:goingaroundocclusion}}
    \vspace{-0.1in}
\end{figure}

\begin{table}
{\footnotesize
    \begin{center}
    \begin{tabular}{|r|l|l|l|l|}
     \hline
        & \multicolumn{2}{c|}{seen} & \multicolumn{2}{c|}{unseen} \\
      \hline
           & \thead{distance \\ mean, std.} &  \thead{improvement \\ mean, std.} &  \thead{dist. \\ mean, std.} & \thead{improvement \\ mean, std.}  \\
           \hline \hline
           
      random actions & 29.4  $\pm$ 3.2 &  -0.85 $\pm$  2.3 & N/A & N/A \\
       \hline \hline
       \hline
      DNA with distance metric as in \cite{foresight} & 25.9 $\pm$ 12.2 & 5.2$\pm$ 13.7 & 24.6$\pm$10.6 & 2.1$\pm$12.4 \\
       \hline
      DNA with our exp. dist. metric eqn. \ref{eq:cost} & \textbf{9.2 $\pm$7.4} &  \textbf{15. $\pm$  11.5} & \textbf{17.5 $\pm$ 10.2} &  \textbf{8.3 $\pm$ 11.8}\\ 
      \hline
      SNA exp. dist. metric eqn. \ref{eq:cost} (ours)& 13.0 $\pm$  5.6 &  12.4 $\pm$ 8.8 & 18.18 $\pm$ 9.5 & 7.7 $\pm$ 10.5\\
      \hline
    \end{tabular}
    \end{center}
    }
    \caption{Results of the pushing benchmark on 20 different object/goal configurations. Units are pixels in the 64x64 images.}
    \label{table:res_longd}
    \vspace{-0.2in}
\end{table}
To examine how well each approach can handle occlusions, we devised a second task that requires the robot to push one object, while keeping another object stationary. When the stationary object is in the way, the robot must move the goal object around it, as shown in \autoref{fig:goingaroundocclusion} on the left. While doing this, the gripper may occlude the stationary object, and the task can only be performed successfully if the model can make accurate predictions through this occlusion. These tasks are specified by picking one pixel on the target object, and one on the obstacle. The obstacle is commanded to remain stationary, while the target object destination location is chosen on the other side of the obstacle.

We used four different object arrangements, with two training objects and two objects that were unseen during training. We found that, in most of the cases, the SNA model was able to find a valid trajectory, while the prior DNA model was mostly unable to find a solution. \autoref{fig:goingaroundocclusion} shows an example of the SNA model successfully predicting the position of the obstacle through an occlusion and finding a trajectory that avoids the obstacle. These findings are reflected by our quantitative results shown in \autoref{table:mult_obj}, indicating the importance of temporal skip connections.

\begin{table}
\centering
{\footnotesize
\begin{tabular}{|r|l|l|l|l|}
 \hline
    & \multicolumn{2}{c|}{seen} & \multicolumn{2}{c|}{unseen} \\
  \hline
        &  \thead{moved imp. \\ mean, std.} &   \thead{stationary imp. \\ mean, std.}  &  \thead{moved imp. \\ mean, std.} &   \thead{stationary imp. \\ mean, std.}  \\
   \hline \hline
  Dynamic Neural Advection, DNA \cite{foresight} & 3.5 $\pm$4.94 &  -1.4 $\pm$ 0.39 & 0.83 $\pm$1.13 &  -1.1 $\pm$ 0.9\\ 
  \hline
  Skip Con. Neural Advection, SNA (ours) & \textbf{8.16 $\pm$ 2.9} &  \textbf{-0.9 $\pm$0.7} & \textbf{10.6 $\pm$ 3.7} & \textbf{-1.5 $\pm$ 0.9} \\
  \hline
\end{tabular}
}
\caption{Results for multi-objective pushing on 8 object/goal configurations with 2 seen and 2 novel objects. Values indicate improvement in distance from starting position, higher is better. Units are pixels in the 64x64 images.} 
\label{table:mult_obj}
    \vspace{-0.2in}
\end{table}

\section{Discussion and Future Work} 
\label{sec:conclusion}
We showed that visual predictive models trained entirely with videos from random pushing motions can be leveraged to build a model-predictive control scheme that is able to solve a wide range multi-objective pushing tasks in spite of occlusions. We also demonstrated that we can combine both discrete and continuous actions in an action-conditioned video prediction framework to perform more complex behaviors, such as lifting the gripper to move over objects.

Although our method achieves significant improvement over prior work, it does have a number of limitations. The behaviors in our experiments are relatively short. In principle, the visual MPC approach can allow the robot to repeatedly retry the task until it succeeds, but the ability to retry is limited by the model's ability to track the target pixel: the tracking deteriorates over time, and although our model achieves substantially better tracking through occlusions than prior work, repeated occlusions still cause it to lose track. Improving the quality of visual tracking of the designated pixels may allow the system to retry the task until it succeeds. More complex behaviors, such as picking and placing (e.g., to arrange a table setting), may also be difficult to learn with only randomly collected data. We expect that more goal-directed data collection would substantially improve the model's ability to perform complex tasks. Furthermore, better predictive models that incorporate hierarchical structure or reason at variable time scales would further improve the capabilities of visual MPC to carry out temporally extended tasks. Fortunately, as video prediction methods continue to improve, we expect methods such as ours to further improve in their capability.

\section*{Acknowledgments}
We would like to thank Prof. Dr. Patrick van der Smagt from the Technical University of Munich for insightful technical discussions and helping to organize Frederik Ebert's visit at UC Berkeley. We would also like to thank Roberto Calandra for assistance the robotic experiments and 3D printing. Furthermore we would like to thank Ashvin Nair and Pulkit Agrawal for insightful discussions. This work was supported by a fellowship within the FITweltweit program of the German Academic Exchange Service (DAAD), a stipend from the German Industrial Foundation (Stiftung Industrieforschung), support from Siemens, the National Science Foundation through IIS-1614653 and IIS-1651843, and an equipment donation from NVIDIA. Chelsea Finn and Alex X. Lee were also partially supported by National Science Foundation Graduate Research Fellowships. 

\bibliography{references}

\newpage
\section*{Appendix} 

\label{sec:appendix}

\subsection*{Hyperparameters}
The video prediction network was trained on sequences of 15 steps taken from 44,000 trajectories of length 30 by randomly shifting the 15-step long window thus providing a form of data augmentation.
The model was trained on 66,000 iterations with a minibatch-size of 32, using a learing rate of 0.001 with the Adam optimizer \cite{ADAM}.
For visual MPC we use $n_{iter}=3$ CEM-iterations. At every  CEM iteration $M=200$ action sequences are sampled and the best $K=10$ samples are selected, fitting a Gaussian to these examples. Then new actions are sampled according to the fitted distribution and the procedure is repeated for $n_{iter}$ iterations.

\subsection*{Prototyping in Simulation}
The visual MPC algorithm has been prototyped and tested on a simple block pushing task simulated in the MuJoCo physics engine. The code has been made available in the same repository.
Using the simulator, training data could be collected orders of magnitude faster. Furthermore a benchmark was set up which does not require any manual reset or manual labeling of the objects' positions. The main downside of simulation however is that the complexity of both real-world dynamics and real-world visual scenes cannot be matched. (More realistic simulators exist, but require very large computational resources and large amounts of hand-engineering for setup).

\subsection*{Dependence of Model Performance on Amount of Training Data}
\begin{figure}
    \centering
    \begin{subfigure}{1.\textwidth}
    \centering
        \includegraphics[width=1\linewidth]{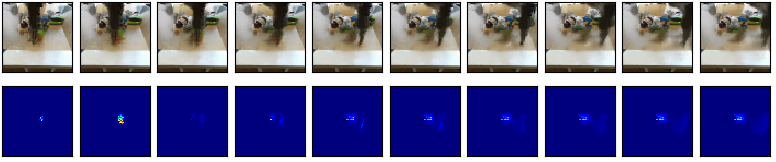}
        \caption{using 5\% of data}
        \label{fig:subfig1}
    \end{subfigure}
    \vskip3mm
    
    \begin{subfigure}{1.\textwidth}
    \centering
        \includegraphics[width=1\linewidth]{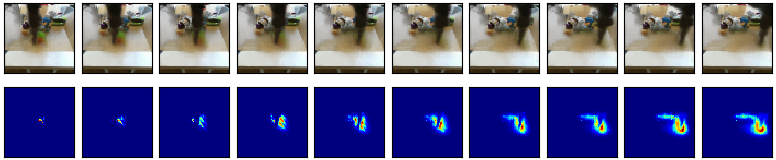}
        \caption{using 10\% of data}
        \label{fig:subfig2}
    \end{subfigure}
    
    \vskip3mm
    
    \begin{subfigure}{1.\textwidth}
    \centering
        \includegraphics[width=1\linewidth]{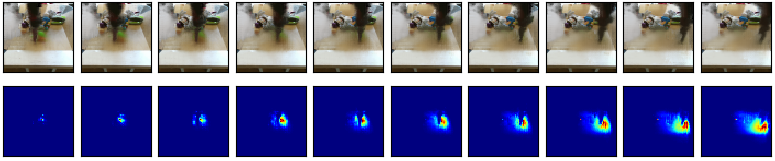}
        \caption{using 100\% of data}
        \label{fig:subfig3}
    \end{subfigure}
    
    \caption{
    Top rows: Predicted images. Bottom rows: Predicted probability distributions of designated pixel on the green object, marked in \autoref{fig:desig_pix_bluedot}. When using only 5\% of the data (of 44k trajectories in total), the model is not able to predict object movement at all (in fact the arm passes through the objects) and the distribution remains on the initial position of the object. For 30\% percent the object distribution appears much more smeared than with 100\% of the data.}
    \label{fig:data_amount}
\end{figure}

We tested the effect of using different amounts of training data and compare the results visually, see \autoref{fig:data_amount}.
The complete size of the data set is 44,000 trajectories. For this evaluation all models were trained for 76k iterations. The quality of the predictions for the probability distribution of the designated pixel vary widely with different amounts of training data and the quality of these predictions is crucial for the performance of visual MPC. The fact that performance still improves with very large amounts of training data suggests that the model has sufficient expressive power to leverage this amount of experience. In the future we will investigate the effect of using even more data and evaluate the effect of a larger number of different training objects.

\subsection*{Failure Cases and Limitations}
Visual MPC fails when objects have vastly different appearance than those used in training. For example when objects are much bigger than any of the training objects these methods tend to perform poorly. Also we observed failure for one object with a very bright color which did not occur in the training set. However the model usually generalizes well to novel objects of similar size and appearance as the training objects. Apart from that we observed two main failure modes: 
The first one occurs when the probability mass of the designated pixel, $P_{t,d^{(i)}}$ drifts away from the object of interest (as discussed in the \autoref{sec:conclusion}) since there is no feedback correcting the position of the designated pixel. To alleviate this problem a tracker can be integrated into the system enabling visual MPC to keep retrying when object behave differently than expected.

The second failure mode occurs when visual MPC does not find an action sequence which moves the designated pixel closer to the goal. This can happen, when the goalpoint for pushing the object is far away and it is very unlikely to sample a trajectory moving closer to the goal. In some cases the problem can be avoided simply by increasing the number of samples used when applying CEM. However this slows down the planning process. In order to enable more temporally extended action sequences a potential solution could be to increase the efficiency of the sampling process e.g. by introducing macro-actions which consist of a sequence of actions (learning macro-actions is out of the scope of this paper). 

\subsection*{Robustness of the Model}

We did not find any negative influence of clutter in the workspace, as long as the object to be moved has a free way path to its goal. When an obstacle is on the path it can be marked by a second designated pixel to avoid collision using an additional cost function. In this case the planner usually manages to find a way around the obstacle. However visual MPC with the current type of model is not yet capable of reasoning by itself that it has to push and object around obstacles when it is not marked explicitly. The reason is that the model has large uncertainty when multiple collisions occur (i.e. when the arm pushes object 1 and object 1 collides with object 2).

Our model is robust to small changes of the viewpoint (in the order of several centimetres in translation and several degrees in orientation). A likely reason is that during data collection the camera has been slightly displaced in position. Robustness could be improved by training a model from several viewpoints at the same time. As the prediction problem also becomes harder in this way more data might be required. 

\subsection*{General skip connection neural advection model}

\begin{figure}
    \centering
    \includegraphics[width=\textwidth]{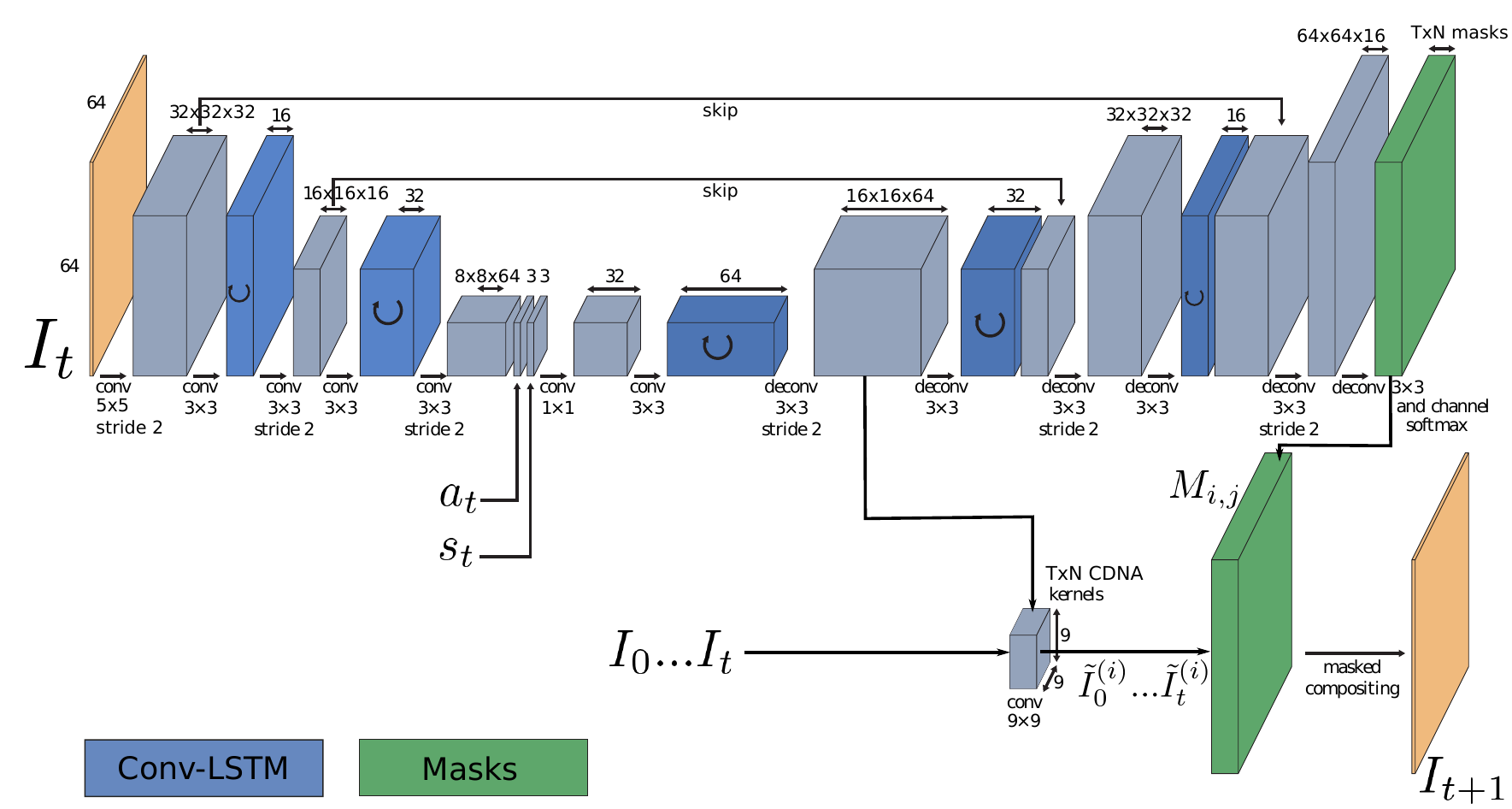}
    \caption{General SNA model based on \autoref{eqn:general_model}.}
    \label{fig:general_model}
\end{figure}

We show a diagram of the general skip connection neural advection model in Figure~\ref{fig:general_model}

\subsection*{Alternative Model Architectures}
Instead of copying pixels from the first image of the sequence $I_0$, the pixels in the first image can also be transformed and then merged together with the transformed pixels from the previous time step as
\begin{equation}
\hat{I}_{t+1} = \tilde{I}_0 \mathbf{M}_{N+1} +  \sum_{i=1}^{N} \tilde{I}_t^{(i)} \mathbf{M}_i
\end{equation}
where $\tilde{I}_0$ is the transformed background. We tested this model on various pushing task and found that performance is comparable to the SNA model.

\end{document}